\documentclass[10pt,journal,cspaper,compsoc]{IEEEtran}
\usepackage{epsfig}
\usepackage{graphicx}
\usepackage{amsmath}
\usepackage{amssymb}
\usepackage[tight]{subfigure}
\usepackage{color}
\usepackage{url}
\usepackage{cite}
\usepackage{blindtext}
\usepackage[switch,displaymath, mathlines]{lineno}
\usepackage{booktabs}
\usepackage{multirow}

\newcommand{\M}{\mathcal{M}}

\newcommand{\trace}{\text{trace}}
\newcommand{\cov}{\text{cov}}
\newcommand{\dist}{\text{dist}}
\newcommand{\s}{\mathbf{s}}

    \newcommand*\patchAmsMathEnvironmentForLineno[1]{%
      \expandafter\let\csname old#1\expandafter\endcsname\csname #1\endcsname
      \expandafter\let\csname oldend#1\expandafter\endcsname\csname end#1\endcsname
      \renewenvironment{#1}%
         {\linenomath\csname old#1\endcsname}%
         {\csname oldend#1\endcsname\endlinenomath}}%
    \newcommand*\patchBothAmsMathEnvironmentsForLineno[1]{%
      \patchAmsMathEnvironmentForLineno{#1}%
      \patchAmsMathEnvironmentForLineno{#1*}}%
    \AtBeginDocument{%
    \patchBothAmsMathEnvironmentsForLineno{equation}%
    \patchBothAmsMathEnvironmentsForLineno{align}%
    \patchBothAmsMathEnvironmentsForLineno{flalign}%
    \patchBothAmsMathEnvironmentsForLineno{alignat}%
    \patchBothAmsMathEnvironmentsForLineno{gather}%
    \patchBothAmsMathEnvironmentsForLineno{multline}%
    }

\ifCLASSOPTIONcompsoc
\else
\fi
\ifCLASSINFOpdf
\else
\fi
\linenumbers
\begin{document}
%
\title{A Diffusion Process on Riemannian Manifold for Visual Tracking}

%
%
%
%

\author{Marcus Chen,
        Cham Tat Jen,
        Pang Sze Kim,
        Alvina Goh 
\IEEEcompsocitemizethanks{\IEEEcompsocthanksitem Marcus Chen and Cham Tat Jen are with
 the Department of School of Computer Engineering, Nanyang Technological University, is with the Department.
\IEEEcompsocthanksitem Pang Sze Kim and Alvina Goh are with DSO National Laboratories, Singapore.}
\thanks{}}

%
%

\markboth{}
{Shell \MakeLowercase{\textit{et al.}}: Bare Demo of IEEEtran.cls for Computer Society Journals}
%


\IEEEcompsoctitleabstractindextext{%
\begin{abstract}
Robust visual tracking for long video sequences is a research area that has many important applications.  The main challenges include how the target image can be modeled and how this model can be updated. In this paper, we model the target using a covariance descriptor, as this descriptor is robust to problems such as pixel-pixel misalignment, pose and illumination changes, that commonly occur in visual tracking. We model the changes in the template using a generative process. We introduce a new dynamical model for the template update using a random walk on the Riemannian manifold where the covariance descriptors lie in. This is done using  log-transformed space of the manifold to free the constraints imposed inherently by positive semidefinite matrices. Modeling template variations and poses kinetics together in the state space enables us to jointly quantify the uncertainties relating to the kinematic states and the template in a principled way. Finally, the sequential inference of the posterior distribution of the kinematic states and the template is done using a particle filter. Our results shows that this principled approach can be robust to changes in illumination, poses and spatial affine transformation. In the experiments, our method outperformed the current state-of-the-art algorithm - the incremental Principal Component Analysis method\cite{Ross08incrementallearning}, particularly when a target underwent fast poses changes and also maintained a comparable performance in stable target tracking cases.
\end{abstract}

}

\maketitle

\IEEEdisplaynotcompsoctitleabstractindextext

%
\IEEEpeerreviewmaketitle

\section{Introduction}
\label{sec:Introduction}
\begin{figure}[!htp]
\centering
\subfigure{
\includegraphics[width=0.45\textwidth]{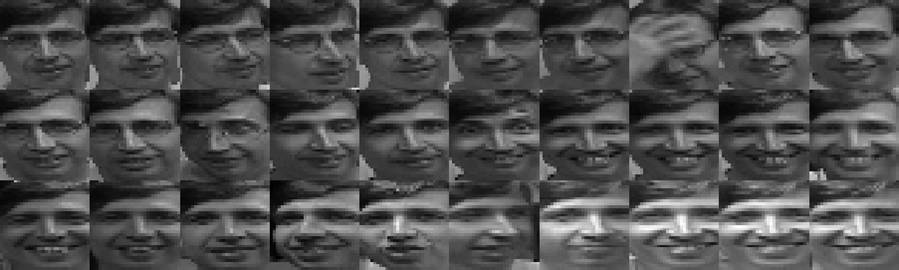}}
\subfigure{
\includegraphics[width=0.45\textwidth]{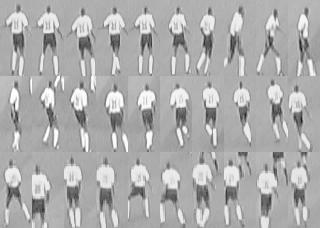}}
\caption{\label{fig:why_update_template}Target patches for successive 871 frames, from \#1, 31, ...871 from 2 video sequences. Target changes in both illumination, poses, appearances even after being affine warped to a standard size.}
\end{figure}

Visual tracking is an important vision research topic that has many applications, ranging from motion-based recognition \cite{cedras1995motion}, surveillance \cite{javed2006tracking}, human-computer interaction \cite{cowie2001emotion}, etc. It also covers many aspects of computer vision problems, such as target feature representation\cite{Yilmaz06objecttracking}, feature selection \cite{collins2005online}, and feature learning \cite{grabner2007learning}.
Even though it has been actively researched for decades, many challenges remain especially with changes in target poses and appearance, and illumination in a long video sequence. Figure \ref{fig:why_update_template} shows two simple examples of how a target can vary over a short time interval. Often these challenges are common and require a good solution in order for long stable tracking in many real life tasks.
There are generally three common approaches to deal with target appearance variations. First is to use robust or invariant target features such as scale invariant feature transformation and color histogram \cite{Birchfield98ellipticalhead}. However, as shown by Figure \ref{fig:why_update_template}, target appearance can change significantly over time, and end up totally different from the starting frame due to variations in target poses and image illumination. The second approach is to employ a complete set of possible target models \cite{black1998eigentracking}, aiming to model possible target variations. However, this requires learning of the target model in advance and can hardly be scalable.  Finally, the last approach is to update the template gradually as it evolves. Note that in this paper, we loosely use the term template for target representation, and do not strictly limit to the image patches. There are several choices for a target template found in the literature.  For  example,  \cite{suard2006pedestrian}  uses  the  histogram  of  oriented gradients, while \cite{Birchfield98ellipticalhead}  uses the color histogram, \cite{sparse_wu2011visual} L1 sparse representation, \cite{Matthews04thetemplate} active appearance model, \cite{Ross08incrementallearning} principal subspace of image patches, and \cite{Porikli:CVPR2006} features covariance.

The template update problem can be expressed mathematically as Eqn. \eqref{eq:heuristic model update}.
 \begin{align}\label{eq:heuristic model update}
  \overline{T_t} = f \left(T_t, \overline{T_{t-1}} \right)
\end{align}
where  $T_t, \overline{T_t}, t\in[1,2,...]$ are the estimated  and updated templates respectively at time $t$. However, as shown in \cite{Matthews04thetemplate}, target template updating is a challenging task.
According to \cite{Matthews04thetemplate}, if the template was not updated at all, the template would become outdated shortly and cannot be used for matching as the target appearance would have undergone changes temporally.  On the other hand, update at every frame would result in accumulation of small errors, and eventually a template drift and loss target information.

Recognizing the importance of template update, many methods have been proposed. One common and intuitive approach is to use linear updating function in the respective feature spaces, such as \cite{Porikli:CVPR2006} on the covariance manifold. This will smoothen the changes between the estimated Template and updated template. Similarly, Kalman filter has also been used in \cite{nguyen2001occlusion} to track template features variables, but not target trajectory. On the other hand, there are three well-known template update algorithms in the literature, namely template alignment \cite{Matthews04thetemplate}, Online Expectation and Maximization (EM) \cite{Jepson@PAMI2003}, and incremental subspace method \cite{Ross08incrementallearning}. Here, we briefly survey these three algorithms.

In template alignment method, \cite{Matthews04thetemplate} proposes a heuristic but robust criteria to decide whether to update the template at time $t$. The basic idea is to keep the starting template to correct the drift of the estimated template. The latest estimated template is first matched to the previous updated template. It is then warped before checking with the first template. For a small template displacement, this method works very well. However, by imposing alignment between the latest template and the first template, this method inherently limit target poses changes to a warping model.

The online EM method \cite{Jepson@PAMI2003} employs a mixture of three template distributions to account for template variations, namely, \emph{long term stable template}, \emph{interframe variational template}, and \emph{outlier template}.
These templates model stable appearance of target, interframe changes in appearance of poses, and occlusion or outliers respectively. Employing a Gaussian mixture model, parameters and membership are estimated on the fly using online EM. In this framework, each pixel in the target patches is assumed to be independent and consequently more stable pixels tend to gain more weights in the similarity measure. This could gradually drift the template in the presence of more stable background pixels.

\begin{figure}[!htp]
\centering \vspace{-10pt}
\subfigure[Representative eigenbases]{
\includegraphics[width=0.15\textwidth]{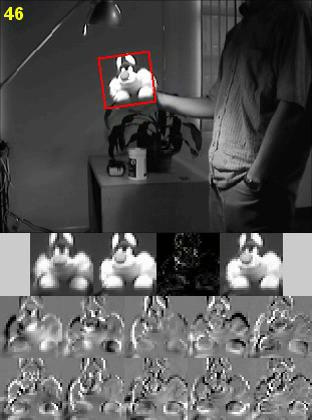}}
\subfigure[Eigenbases of misaligned target regions]{
\includegraphics[width=0.15\textwidth]{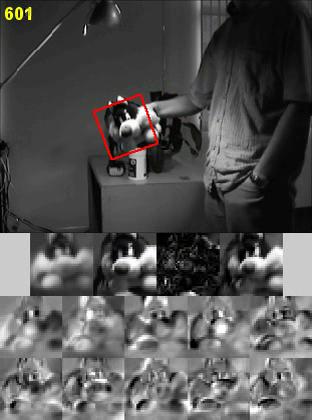}}
\subfigure[Eigenbases are not representative]{
\includegraphics[width=0.15\textwidth]{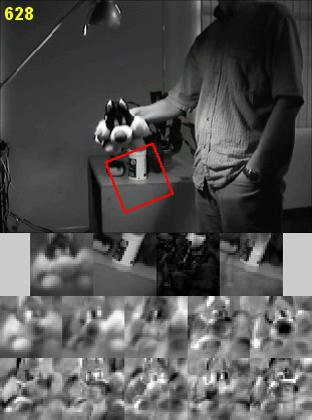}}
\caption{\label{fig:eigen basis}Results of incremental subspace method on the Sylvester sequence. Pixel-wise misalignment could render eigenbasis non-representative. The $1^{st}$ row are the sample frames. The $2^{nd}$ row images are the current sample mean, tracked region, reconstructed image, and the reconstruction error respectively. The $3^{rd}$ and $4^{th}$ rows are the top $10$ principal eigenbases.}
\end{figure}

The third algorithm is to represent the target in its eigenspace, proposed by \cite{Ross08incrementallearning}. The posterior estimates of the template are collected over an interval, and these estimates are then analyzed online through an Incremental Principal Component Analysis method(IPCA). This method can capture changes in template variation in eigenbases. The mean of the posterior estimates are also kept as stable templates. The authors have tested IPCA with various video sequences, and demonstrated its great robustness to the template variations due to pose changes and illumination changes.  Figure \ref{fig:eigen basis} illustrates an incremental update of eigenbases and means. The images in the $3^{rd}$ row show how the eigenbases evolve over time. It has been shown in the paper that the updated templates could almost reconstruct the original image samples over the sequence, reflecting the ability of the eigenbases to model temporal variations.
%
%
Although IPCA is often very robust and can track target very accurately even in noisy, low contrast image sequences, IPCA falls short when the target undergoes fast pose changes and dramatic illumination changes as stated on the paper. This may be because PCA inherently assumes that the target templates over time are from a Gaussian distribution. In abrupt changes in poses and illumination, this assumption does not hold. The unimodal distribution also requires good pixel-wise alignment between the posterior estimate and eigenbases, otherwise uncertainties in template alignment would contribute to template variance and may lead to non-informative basis. A good example from the paper is shown in Figure \ref{fig:eigen basis}. One can see that from frames $\#600$ to $\#636$, the eigenbases are not representative anymore and the tracker loses track of the target.
\vspace{-10pt}
\medskip

So far, most of the current state-of-the-art algorithms update templates in an out-of-chain manner, by assuming the posterior estimate is ``good enough" for template update with pixel-wise alignment. If the target poses posterior estimate is inaccurate or there is a mis-alignment between the estimated and last updated template, the update methods will gradually drift. On the other hand, if the template update is not good, then the posterior estimate of target poses is unlikely to be accurate. These coupled dual problems often render these methods unable to track well when the targets undergo fast changes in poses or non-rigid transformation. However, robustness to fast target poses has many real life applications such as human tracking, maritime target tracking, etc.

To solve these dual problems faced by the existing state-of-the-art algorithms, \cite{chen2011visual} introduces a novel approach to simultaneously quantify these two uncertainties by including both of them into the state space of a Bayesian framework, instead of just target poses in the exist methods. In this manner, no posterior estimate is used for updating, instead better matched multiple hypothesized templates are propagated automatically.

\noindent\textbf{Paper contributions.}
To the best of our knowledge, almost all the state-of-art algorithms use out-of-chain template updating methods. That is to say, the updating of template model is done after obtaining the posterior estimate of the targets position. In this paper, we propose a method to update target model in tandem with the target kinematics. In other words, we model the target template as a part of the state space. We
choose the covariance descriptor for the target descriptor as it is more robust to problems such as pixel-pixel misalignment and changes in pose and illumination. Since positive definite covariance matrices form a Riemannian manifold, we model the target template model variation by a random walk on
the covariance Riemannian manifold. We propose a novel superior template propagation mechanism in the log-transformed space of the manifold to free the constraints imposed inherently by positive semidefinite matrices, leading to a greater ability in dealing with template variations. Our resultant method outperforms the state-of-the-art Incremental PCA algorithm \cite{Ross08incrementallearning} in dealing with fast moving and changing targets, as will be clearly shown in the experiments section.

The  paper  is  organized  as  follows: Section \ref{section:related work and motivation} gives a brief introduction to both covariance descriptors and Riemannian manifold, Section \ref{section:Bayesian Framework} gives a Bayesian formulation of simultaneous inference of both target kinetics and template posterior distribution, Section \ref{section: Analysis of Random Walk} analyzes the template generative process. In Section \ref{section:experiments}, we empirically compare our results with IPCA and give a short discussion. Finally, section \ref{section:conclusion} concludes this paper.

\section{Target Covariance Descriptor}
\label{section:related work and motivation}

In this section, we explain the motivation of using covariance descriptor and its operation on Riemannian manifold.
\subsection{Covariance Descriptor}
\label{subsection:Covariance Descriptor}
A covariance descriptor is defined as follows:
\begin{equation}\label{CovFormula}
    C= \frac{1}{N-1} \sum_{i=1}^N \left(f(i)- \bar{f} \right) \left(f(i)-  \bar{f} \right)^T
\end{equation}
where $f$ is a feature vector, $\bar{f} = {1 \over N}\sum_{i=1}^N (f(i))$ is the mean of the feature vector over $N$ pixels in the target region. In this paper, we use the following $9$-dimensional feature vector:
\begin{align}\hspace{-10pt}
\nonumber f(i)= \left[ x_w,y_w, I(x_w,y_w), |I_{x_w}|,|I_{y_w}|, \sqrt{I_{x_w}^2+I_{y_w}^2}, \right.
\\ \left.  \arctan{ \frac{|I_{x_w}|}{|I_{y_w} |},|I_{xx_w} |,|I_{yy_w}|}\right].
\end{align}
They are $x,y$ coordinates, pixel intensity, $x, y$ directional intensity gradients, gradient magnitude and angle,  and second order gradients respectively. $w$ denotes that these features are extracted after warping image patches to a standard size.

Since its proposed use in human detection \cite{tuzel2006region}, covariance descriptor has gained popularity for many applications, such as face recognition \cite{pang2008gabor}, license plate detection \cite{porikli2006robust}, and tracking \cite{Porikli:CVPR2006, sparse_wu2011visual}.
Some main advantages of choosing the covariance descriptor \cite{Tuzel:CVPR2007} to model the template  include its lower dimensionality of  ${1\over 2}(d^2+d)$ ($45$ in this paper as $d=9$), compared to its number of target pixels ($32\times32 = 1024$ in this paper), its ability to fuse multiple possibly correlated features, and its robustness to match targets in different views and poses.

By its definition, covariance matrix is clearly a positive semi-definite matrix, which lies on a Riemannian manifold. We will now briefly explain some basic operations on the Riemannian manifold.
\subsection{Riemannian Manifold}
\label{subsection:Riemannian manifold}
\begin{figure}[!htb]
\centering
  \includegraphics[width=0.25\textwidth]{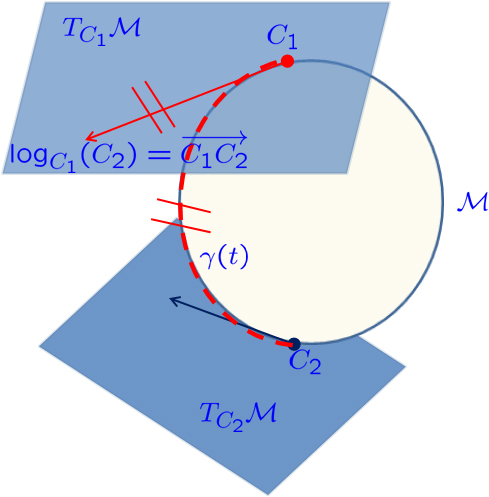}
  \caption{\label{fig:riemannianIllustration1}The geodesic distance is the norm of a vector on the tangent space $T_{C_1}\M$ of at point $C_1$ on the manifold $\M$}
\end{figure}
\begin{table}[!htb]
 \caption{Operations in Euclidean and Riemannian spaces}
 \label{table:corresponding addition and subtraction on Riemannian manifold}
\centering
    \begin{tabular}{ll}
      \hline
       Euclidean space & Riemannian manifold  \\  \hline
       $ C_j = C_i+\overrightarrow{C_i C_j}$ & $C_j =\exp_{C_i}(\overrightarrow{C_i C_j})$ \\
      $\overrightarrow{C_i C_j}= C_j-C_i$ & $\overrightarrow{C_i C_j} = \log_{C_i}(C_j)$\\
      $\dist(C_i,C-j) = \|C_j-C_i \|$  & $\dist(C_i,C_j) = \|\overrightarrow{C_i C_j}\|_{C_i}$\\
      \hline
    \end{tabular}
 \end{table}
A Riemannian manifold $\M$ is a differential manifold and each of its tangent space $T_{C_i}\M$ has a metric function $g$ which defines the dot products between any two tangent vectors $y_k, y_l$. The covariance descriptor is a point on the manifold $\M$, the following operations can be applied to it.
The Riemannian metric:
\begin{align}
\label{yzDistance}
  \langle y_k,y_l\rangle_{C_i}=\trace \left(C_i^\frac{1}{2} y_k C_i^{-1} y_l C_i^{-\frac{1}{2}} \right).
\end{align}
The exponential map $\exp_{C_i}: T_{C_i}\M\rightarrow \M$, takes a tangent vector at point $C_i$ and maps to another point $C_j$: \begin{align}
\label{riemannian exponential}
C_j=\exp_{C_i}(y)={C_i}^{1/2}  \exp \left({C_i}^{-\frac{1}{2}} y {C_i}^{-\frac{1}{2}} \right) {C_i}^{\frac{1}{2}},
\end{align}
The inverse of the exponential map is the logarithm map, which takes a starting point $C_i$ and destination $C_j$, maps to the tangent vector $y$ at point $C_i$.
\begin{align}
\label{yzDistance}
 y= \log_{C_i}{{C_j}}= {C_i}^\frac{1}{2}\log \left({{C_i}^{-\frac{1}{2}} {C_j}{C_i}^{-\frac{1}{2}} }\right) {C_i}^\frac{1}{2}.
\end{align}
Finally, the distance between two covariance matrices ${C_i}$ and $C_j$ is given as:
\begin{align}
\label{eq:distCov}
    d(C_i,C_j)= \sqrt{\sum_{k=1}^{d} \ln^2 \lambda_k \left(C_i,C_j \right)},
\end{align}
where $\lambda_k \left(C_i,C_j \right)$ are the generalized eigenvalues of $C_i$ and $C_j$. That is, $\lambda_k C_i v_k-C_j v_k=0$, and $d$ is  the dimension of the covariance matrices.

Note that $\exp_{C_i}(\cdot)$ and $\log_{C_i}(\cdot)$ are maps on the Riemannian manifold, whereas $\exp(\cdot)$ and $\log(\cdot)$ denote the normal matrix exponential and logarithmic operations. Both $\exp_{C_i}(y)$ and tangent vector $y$ are both $d \times d$ matrices in this paper.

\subsection{Motivation of Manifold Modeling}
\begin{figure}[!t]
\centering
  \includegraphics[width=0.25\textwidth]{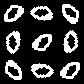}
  \caption{\label{fig:rotated_zeros}A collection of rotated handwriting zeros by the angle of $45^0$ each time.}
\end{figure}
\begin{figure}[!t]
\centering
\subfigure[Targets]{
  \includegraphics[width=0.18\textwidth]{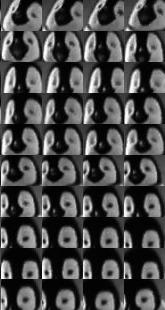}}
  \subfigure[Backgrounds]{
  \includegraphics[width=0.18\textwidth]{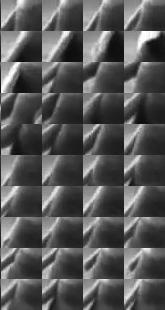}}
\subfigure[SVM results]{
  \includegraphics[width=0.08\textwidth]{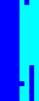}}
  \subfigure[Euclidean distance]{
  \includegraphics[width=0.22\textwidth]{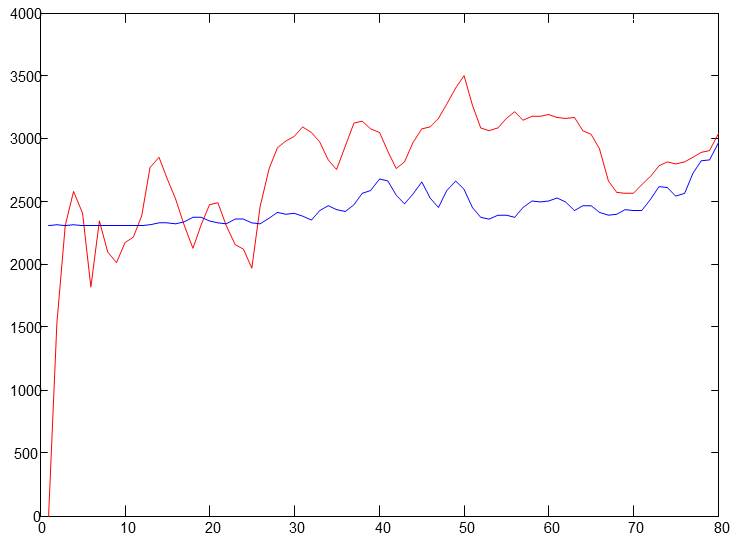}}
    \subfigure[Manifold distance]{
  \includegraphics[width=0.22\textwidth]{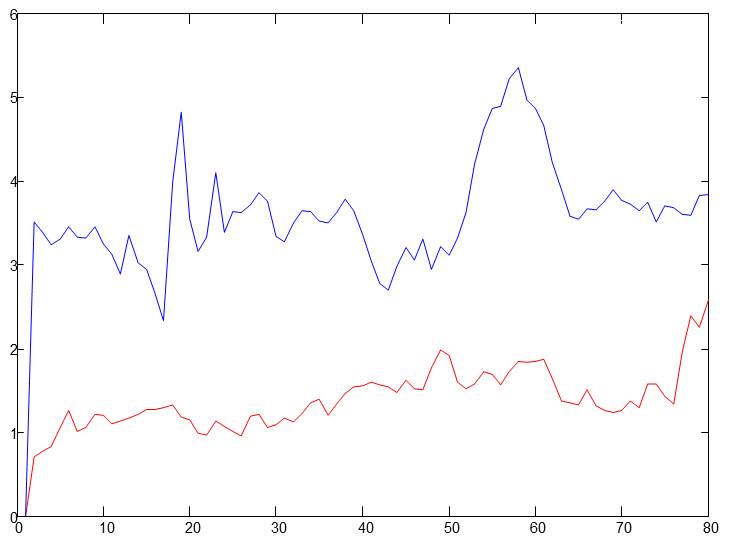}}
  \caption{\label{fig:why manifold}An illustration of distance between images can be better modeled on a manifold, a sequence of dancing penguin from youtube. In (d), Euclidean distance between targets is larger than the distance between target and background; and not on the manifold space. Furthermore, in (e), SVM cannot linearly separate the targets from backgrounds in vectorized Euclidean space.}
\end{figure}

High dimensional image data often lies in low dimensional manifold. For an example, a collection of rotated handwriting zeros in Figure \ref{fig:rotated_zeros} lie in a dimension of $28 \times 28 = 784$ using vectorized representation, but have only one rotational parameter. Popular dimensional reduction methods such as ISOMAP \cite{isomap}, eigenmap\cite{eigenmap}, LLE\cite{lle} model data using manifold structures. In visual tracking,  the target patches in the image sequence are implicitly bounded by the target's degree of freedom captured by images, such as rotation, translation, scaling etc. These implicit parameters modeled using low dimensional manifold could capture image distance. The simplest and yet most popular distance measure between images is Euclidean distance between the vectorized images. A simple example of the head of dancing penguin from youtube is adopted to illustrate in Figure \ref{fig:why manifold} that the manifold of covariance descriptor can model the image distance better and can separate target patches from the background better. Using Euclidean distance, the distance between target patches and first target patch could be larger than the one between the background and the first target patch. Furthermore, a test of support vector machine using linear kernel showed that some background patches are classified into the target patches. On the other hand, the separation between target patches and background patches were separated shown in Figure \ref{fig:why manifold}e.

\section{Bayesian Framework}
\label{section:Bayesian Framework}
In this section, we use a standard Bayesian framework\cite{ristic04} to formulate tracking of both template and kinetics as follows:
\begin{align} \label{eq:posterior}\hspace{-15pt}
  \nonumber  P(C_{t}, \s_{t} | z_{1:t}) &\propto&  P(z_{t}|C_{t}, \s_{t} ) \int P(\s_t, C_t | \s_{t-1},C_{t-1})\\
    \hspace{-15pt} && P(C_{t-1}, \s_{t-1} | z_{1:t-1}) d\s_{t-1} dC_{t-1},
\end{align}
where $z_t$ is the measurement, $\s_{t}$ is the kinetic state variables, $C_{t}$ is the covariance descriptor,  $P(C_{t}, \s_{t} | z_{1:t})$ is the posterior probability of target template and pose given the measurement, $P(z_{t}|C_{t}, \s_{t})$ is the observational model, and $P(\s_t, C_t | \s_{t-1},C_{t-1})$ is the dynamical model. They are further elaborated in the following subsections.
\subsection{Dynamical Model}
\label{subsection: Dynamical Model}
The state space in our paper includes both target kinetic variables $\s_t$ and template covariance descriptor $C_t$. The state variables are defined in Eqns. \eqref{eq:state kinetic} and \eqref{eq: state covariance}, and we would like to estimate them through the Bayesian framework in Eqn.\eqref{eq:posterior}. These state variables are propagated from time $t-1$ to $t$ through a dynamical model $P(\s_t, C_t | \s_{t-1},C_{t-1})$ .
\begin{align} \label{eq:state kinetic}
 \s_t  &= [x_t,y_t,\dot{x_t}, \dot{y_t}, h_t, \theta_t],\\
\nonumber  C_t &= \cov \left( x_w,y_w, I(x_w,y_w), |I_{x_w}|,|I_{y_w}|, \right. \\ &\left. ~~\arctan{ \frac{|I_{y_w}|}{|I_{x_w} |}, }\sqrt{I_{x_w}^2+I_{y_w}^2},|I_{xx_w} |,|I_{yy_w}|\right), \label{eq: state covariance}
\end{align}
where $x_t, y_t$ are the spatial coordinates of the target center position at time $t$, $\dot{x_t}$, $\dot{y_t}$ are the velocities, $h_t$ is the scaling factor, and $\theta_t$ is the orientation. $x_w, y_w$ are the coordinates of a pixel on the standard target patch warped from $x_t, y_t$, $I(x_w,y_w)$ is the pixel intensity and $\{I_{x_w},I_{y_w}\}$ are the patch intensity gradients, $\{ I_{xx_w}, I_{yy_w}\}$ are the second order gradients. Assuming independence between kinetic variables and covariance, we model the joint dynamics as follows:
\begin{align}
P(\s_t, C_t | \s_{t-1},C_{t-1}) &= P(\s_t | \s_{t-1}) P(C_t |C_{t-1}),\\
      \s_{t} &= k(\s_{t-1})+u_t, \label{eq:kinetic dynamic}\\
      C_t &= \exp_{C_{t-1}}( n_{t}). \label{eq:cov dynamic}
\end{align}
%
%
%
$k$ is the kinetic model and we use a near constant velocity linear model $k(\s_{t-1}) = A\s_{t-1}$. $u_t$ is generated with an interacting Gaussian models with a jumping probability of $[0.9, 0.1]$ to model sudden changes in target poses. As for template dynamical model, $n_{t} \in T_{C_i}\M $ is a random process on the tangent plane of manifold $\M$. An example of this could be the Brownian motion process as described by \cite{ito1987brownian}.
%
In this paper, we choose to model the template dynamical model in log-transformed space of the manifold as follows:
\begin{align}
\label{eq:symmetric rand}
   &C_t = \exp (\log(C_{t-1}) + w_t)
  \\\nonumber
      &\text{where } w_t \sim N(0, \Sigma), ~ \Sigma_{i,j} = \Sigma_{j,i} \sim N\left(0, \sigma_{i,j}^2\right)
 \\      &P(\log(C_t))\propto \exp \left(-\frac{1}{2}\sum_{i\le j,i,j\in[1,d]}\frac{w_t(i,j)^2}{\sigma_{i,j}^2} \right)
\end{align}
where $w_t$ is simply a random symmetric matrices and $N(0, \sigma_{i,j}^2), i,j \in[1,d]$ are normal distributions. According to \cite{arsigny2008geometric}, the matrix exponential function maps a symmetric matrix to its corresponding positive semi-definite, $exp:Sym(d)\rightarrow Sym^+(d)$, and it is one-to one mapping. As such, the generated samples of $C_t$ is always a positive semi-definite (PSD)  matrix. This frees the inherent constraints of positive eigenvalues in a PSD matrix. This distribution may be considered as a log-normal distribution of the PSD matrices as defined in \cite{schwartzman2006random}.

%
\begin{align}
\begin{split}
    &~~C_t^{-1}C_{t-1}\\
    &=\exp \left(-\log(C_{t-1}) - w_t\right)\exp\left(\log(C_{t-1})\right)\\
     &= \exp(-w_t)
\end{split}
\end{align}
Generalized Eigenvalues:

\begin{align}
\begin{split}
\lambda_k C_t v-C_{t-1} v &= 0
\\ C_t^{-1}C_{t-1} v  &= \lambda_k v
\\exp(-w_t)v &= \lambda_k v
\\d(C_{t-1},C_{t}) &= \left[\sum_{k=1}^{d} \left[\ln^2 \lambda_k \left(exp(-w_t)\right)\right]\right]^{1/2}
\\ &=  \left[\sum_{k=1}^{d} \lambda_k^2(w_t)\right]^{1/2}, \, if ~~  \exists w_t^{-1}
\end{split}
\end{align}

In this paper, for $d=9$, $w_t$'s eigenvalues $\lambda_1(w_t)\ge \lambda_2(w_t) \ge ... \ge \lambda_9(w_t)$ can be bounded according to \cite{zhan2006extremal}, assuming the entries of the noise matrix are bounded by $[a, b]$, i.e. $a \le w_t(i,j) \le b$:
 \begin{align}
    \lambda_9(w_t) \ge  \begin{cases}
            \frac{1}{2}\left(9a-\sqrt{a^2+80b^2}\right) & |a|<b \\
            9a & \mbox{otherwise.}
          \end{cases} \\
    \lambda_1(w_t) \le  \begin{cases}
            \frac{1}{2}\left(9b+\sqrt{a^2+80b^2}\right) & |a|<b \\
            9b & \mbox{otherwise.}
          \end{cases}
\end{align}
In other words, the eigenvalues are roughly within an order of magnitude of $\max (\sigma_{i,j})$ for this random process. In this way, the template diffusion spread on the manifold can be easily managed by choosing an appropriate $\max (\sigma_{i,j})$ in $w_t$.

\begin{figure}[!tbp]
\centering
\includegraphics[width=0.48\textwidth]{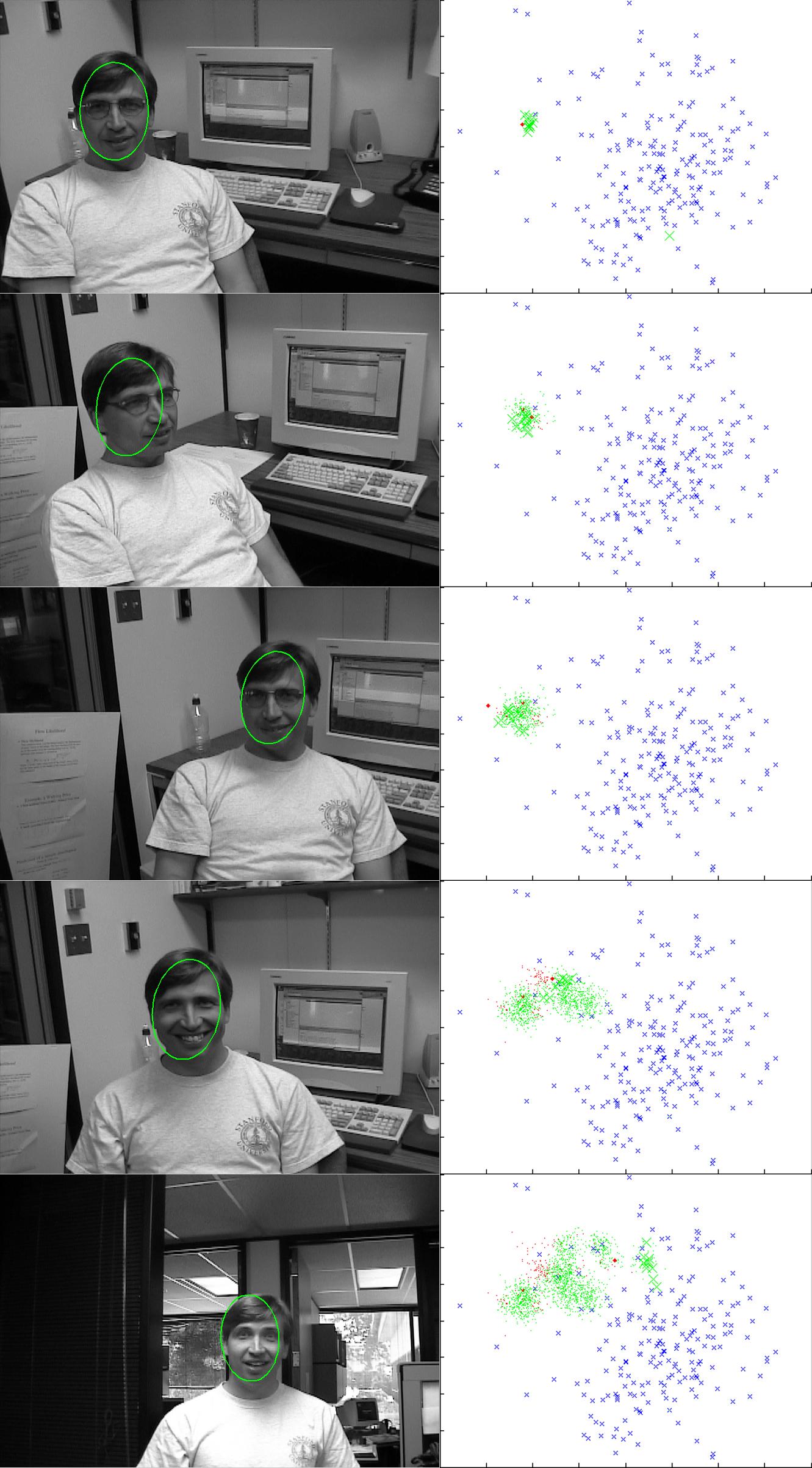}
\caption{Evolution of template and generated templates, frame \#1, 101, 201, 501, 801, Green cross: top 10 similar generated templates, Red: ground truth, Blue cross: the background.}
\label{fig:random walk on the manifold}\vspace{-10pt}
\end{figure}

\subsection{Observation Model}
The observation model $P(z_{t}|C_{t}, \s_{t})$ measures the likelihood of a target given target poses and template values, it is modeled as follows:
\begin{align}
P(z_{t}|C_{t}, \s_{t} )  &\sim  N(0, \sigma^2), \\
\nonumber z_{t} &= d(C_t, C_t^*), \\
\nonumber     C_{t}^* &= g(\s_{t}, \text{Image}),\\
\label{eq:measurement model} P(z_{t}|C_{t}, \s_{t} )  &\propto \exp(-\frac{1}{2\sigma^2}d^2)
\end{align}
Here, $d(C_{t}, C_{t}^*)$ is given by  Eqn. \eqref{eq:distCov}. $g$ is the covariance computation operator; $g$ takes the kinetic value $\s_{t}$ of each particle at time $t$, warps the region to a standard size (in this paper, $32\times32$) before computing covariance.
\subsection{Overall Framework}
We use a standard particle filter to do sequential inference. The particle filter \cite{cappe:pf},\cite{ristic04},\cite{Isard:pf} represents the distribution of state variables by a collection of samples and their weights. The advantage of using a particle filter is that it can deal with non-linear system and multi-modal posterior. The algorithm of the particle filter is as follows:
\begin{enumerate}
  \item \textbf{Initialization}.The particle filter is initialized  with  a known realization of target state variables. This includes the target initial state values. Covariance of the target  $C_0$, i.e. initial template is extracted for comparison later.  The parameters of covariance generative process. i.e. template dynamical model are also determined.

  \item \textbf{Propagation}. Each particle is propagated according to the propagation model in Eqns. \eqref{eq:kinetic dynamic} and \eqref{eq:symmetric rand}. Both kinetic variables and template are generated through these random processes.

  \item \textbf{Measure the likelihood}. At each particle $i$, the covariance descriptor $C_t^*(i)$ extracted is compared to its corresponding template $C_t(i)$. The likelihood of the particle is then estimated as given in Eqn. \eqref{eq:measurement model}.

  \item \textbf{Posterior estimation}. The posterior estimate gives the estimate of the current target state, given all its previous information and measurements. This could be maximum a posteriori probability estimate or minimum mean square error estimate (MMSE). In this paper, we use MMSE.
  \item \textbf{Resampling}. To avoid any degeneracies, resampling is conducted to redistribute the weight of particles.

  \item \textbf{Loop}. Repeat the process from step 2 to 5 as time progresses.
\end{enumerate}

\section{Analysis of Template Generative process}
\label{section: Analysis of Random Walk}
\begin{figure}[!htp]
  \centering
  \includegraphics[width=0.45\textwidth]{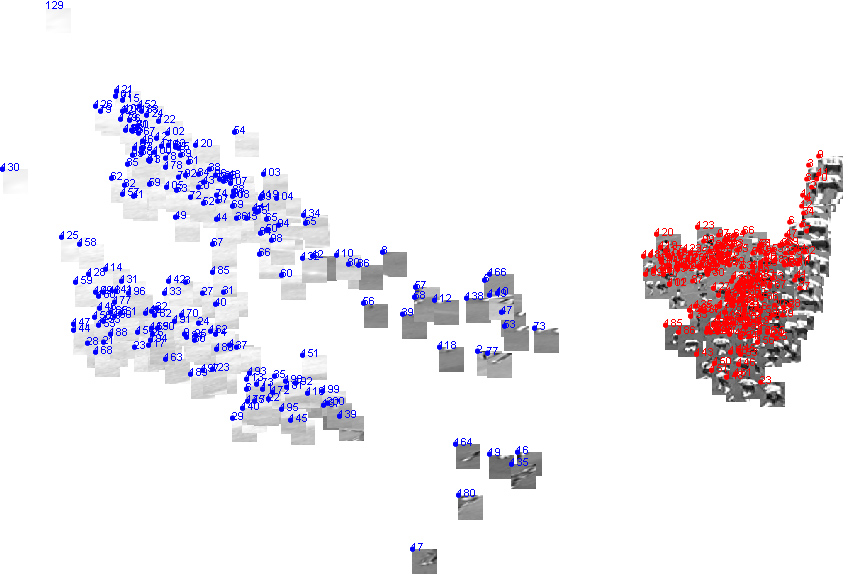}
  \caption{Visualization of target in``soccer" sequences on the covariance manifold, Red:target patches, Blue: background patches.}
  \label{fig:target on manifold separated from background}
\end{figure}
In this section, we show that the covariance descriptor is a good representation of the target as well as the motivation behind performing a random walk as given in Section \ref{subsection: Dynamical Model}. Two reasonable criteria for a good target representation are as follows:
\begin{itemize}
  \item the representation evolves gradually as the target undergoes changes in poses, appearance etc,
  \item there is clear separation of target and background.
\end{itemize}
To help visualize the distribution of target covariance matrices on the manifold, we use multidimensional scaling \cite{mds} to construct a visualization of the distribution of the covariance matrices.  The  distance  matrix  is  constructed  using Riemannian distance as given in Eqn.\eqref{eq:distCov}. The visualization shows the relative positions of targets (red) and backgrounds(blue). Visualizing the PETS 2003 Soccer sequence and Dudek Face sequence in Figures  \ref{fig:target on manifold separated from background} and \ref{fig:random walk on the manifold}, we noticed that our representation of the targets tended to  cluster  together as they evolved gradually. This evolution is smoother and easier to model on the manifold as compared to the evolution of its original feature values at each pixel. This observation  motivated us to model the template variations by using a random walk on the Riemannian manifold. Based  on  Eqns.\eqref{eq:kinetic dynamic} and \eqref{eq:symmetric rand}, Figure \ref{fig:random walk on the manifold} illustrates a realization of the random walk. This shows that our template dynamical model can model the actual target appearance variations. Changes in facial expression and face poses cause covariance template (shown as red points) to evolve slowly on the manifold, and they are well modeled by the generated covariances on the manifold (shown as green points).
\section{Experiments and Results}
\label{section:experiments}
\subsection{Experimental data}
We tested our algorithm on some of popular tracking datasets, David  Ross's  sequences  including
plush toy (toy Sylv), toy dog, david, car 4sequences from his website, Dudek Face sequences, and vehicle tracking  sequences  from  PETS2001,  soccer  sequence  from PETS2003. The test data information is tabulated in Table \ref{tab:track info}.
%
\begin{table*}[!htbp]
  \centering
    \caption{Test Sequences}
    \begin{tabular}{cccc}
    \toprule
    Test sequences & Source & No. of frames & Characteristics \\
    \midrule
    Plush Toy (Toy Sylv) & David Ross & 1344  & fast changing, 3D Rotation, Scaling,  Clutter, large movement \\
    Toy dog & David Ross & 1390  & fast changing, 3D Rotation, Scaling,  Clutter, large movement \\
    Soccer player 1 & \multirow{4}[0]{*}{PETS 2003} & \multirow{4}[0]{*}{1000} & \multirow{3}[0]{*}{Fast changing, white team, good contrast with background, occlusion } \\
    Soccer player 2 &       &       &  \\
    Soccer player 3 &       &       &  \\
    Soccer player 4 &       &       & Fast changing, gray(red) team,poor contrast with background, occlusion  \\
    Dudek Face Sequence & A.D. Jepson & 1145  & Relatively stable, occlusion, 3D rotation \\
    Truck & PETS 2001 & 200   & relatively stable, scaling \\
    David  & David Ross & 503   & relatively stable 2D rotation \\
    Car 4 & David Ross & 640   & Relatively stable, scaling, shadow, specular effects \\
    \bottomrule
    \end{tabular}%
  \label{tab:track info}%
\end{table*}%

\subsection{Performance measure}
As spelled out in \cite{manohar2006performance}, a good measure should include both overall tracking and goodness of track. This paper uses the ratio between on-track length and sequence length to capture the performance of overall tracking, and on-track accuracy for goodness of track. Define tracking errors as: $e_x(t) = \|g_x(t) - x(t)\|, e_y(t) = \|g_y(t) - y(t)\|$, where $e_x(t), e_y(t), g_x(t), g_y(t)$ are the errors in $x,y$ and ground truth in $x,y$ at time $t$ respectively.
\begin{align}
&\gamma_{on track} = \frac{1}{2}\left(\frac{e_x(t)}{H_x(t)} + \frac{e_y(t)}{H_y(t)}\right)\le 1\\
&r_{on track} = \frac{\gamma_{on track}}{l} \\
&rms_{on track} = \sqrt{\left( \frac{e_x(t)}{H_x(t)}\right)^{1/2} + \left(\frac{e_y(t)}{H_y(t)}\right)^{1/2}}
\end{align}
$H_x(t), H_y(t)$ are the ground truth target size at time $t$. In this work, ground truth on the target center is manually annotated, the target size is assumed to as those of the first frame (this may not be applicable to frames with a large change in target size).
\subsection{Results and discussion}
\begin{figure*}[!t]
\centering
\subfigure[Track duration rate $r_{on track}$ ]{
\includegraphics[width=0.36\textwidth]{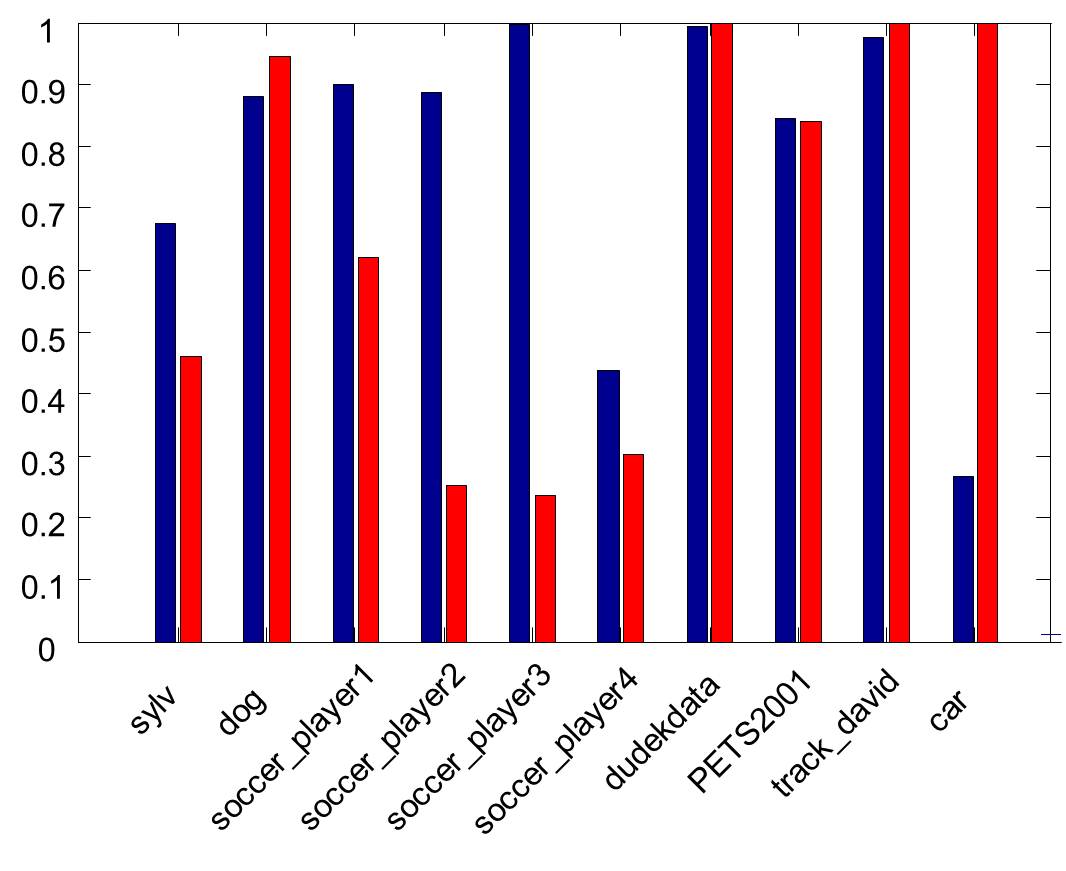}
\label{subfig: Track duration}}
\subfigure[Track accuracy $rms_{on track}$]{
\includegraphics[width=0.6\textwidth]{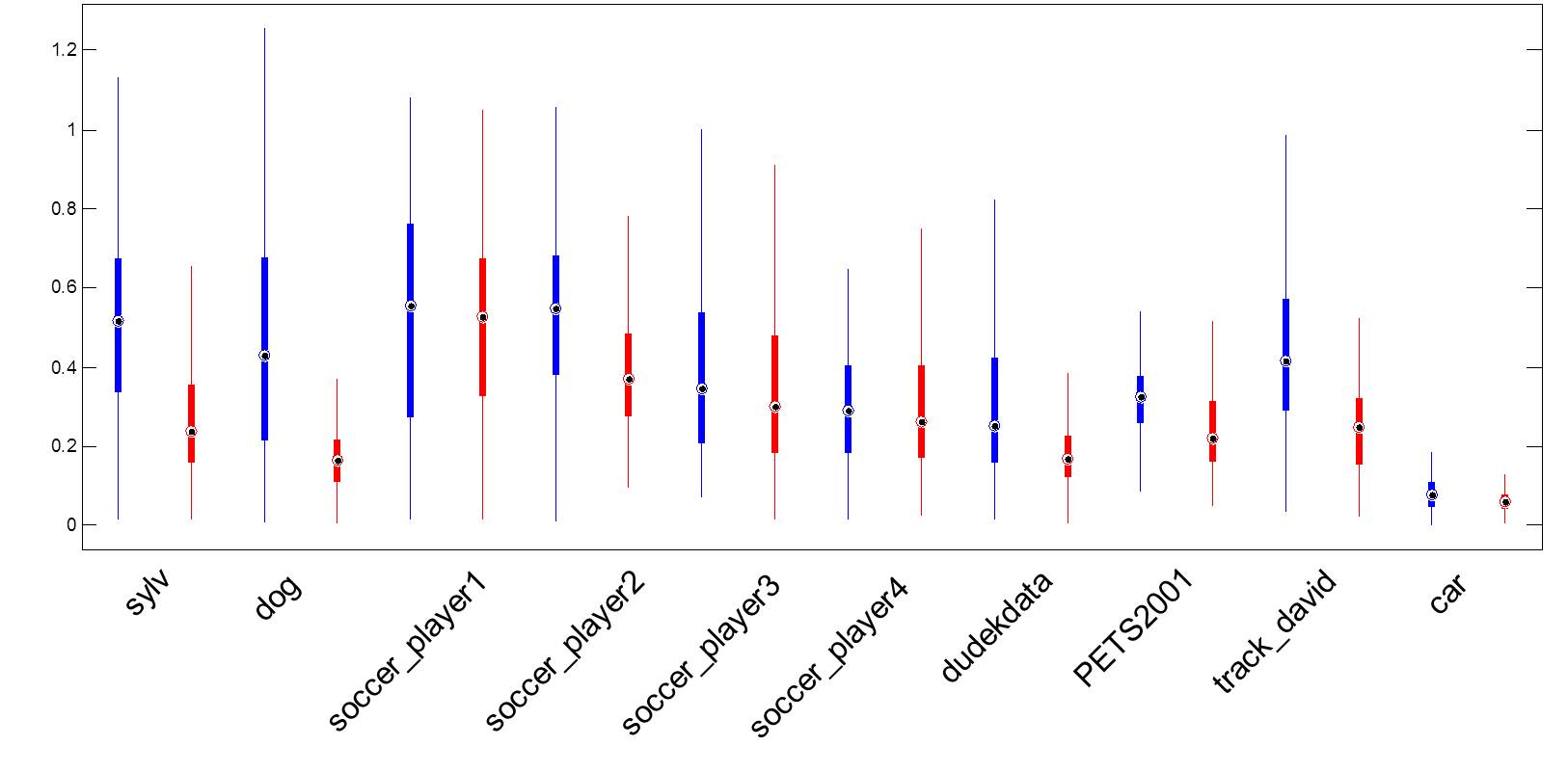}
\label{subfig:Track accuracy}}
\caption{\label{fig:results_stats} The results statistics, our results in blue, IPCA in red.}
\end{figure*}
\begin{figure}[!t]
\centering
\includegraphics[width=0.48\textwidth]{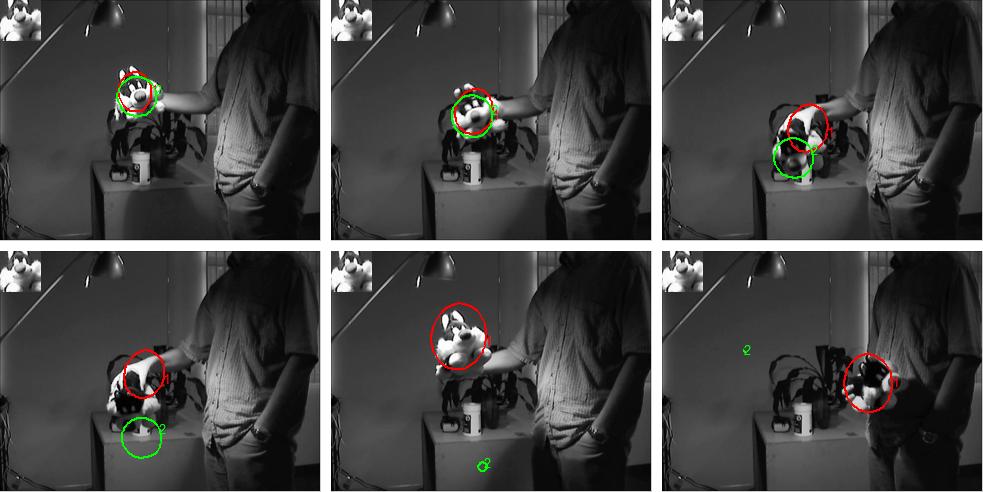}
\caption{ Tracking results on PLUSH TOY SYLV sequences, frame \#133, 594, 609, 613, 957, 1338, Green: IPCA, Red: our results. The IPCA failed to recover track from frame \# 609.}
\label{fig:res_sylv}\vspace{-10pt}
\end{figure}
\begin{figure}[!t]
\centering
\includegraphics[width=0.48\textwidth]{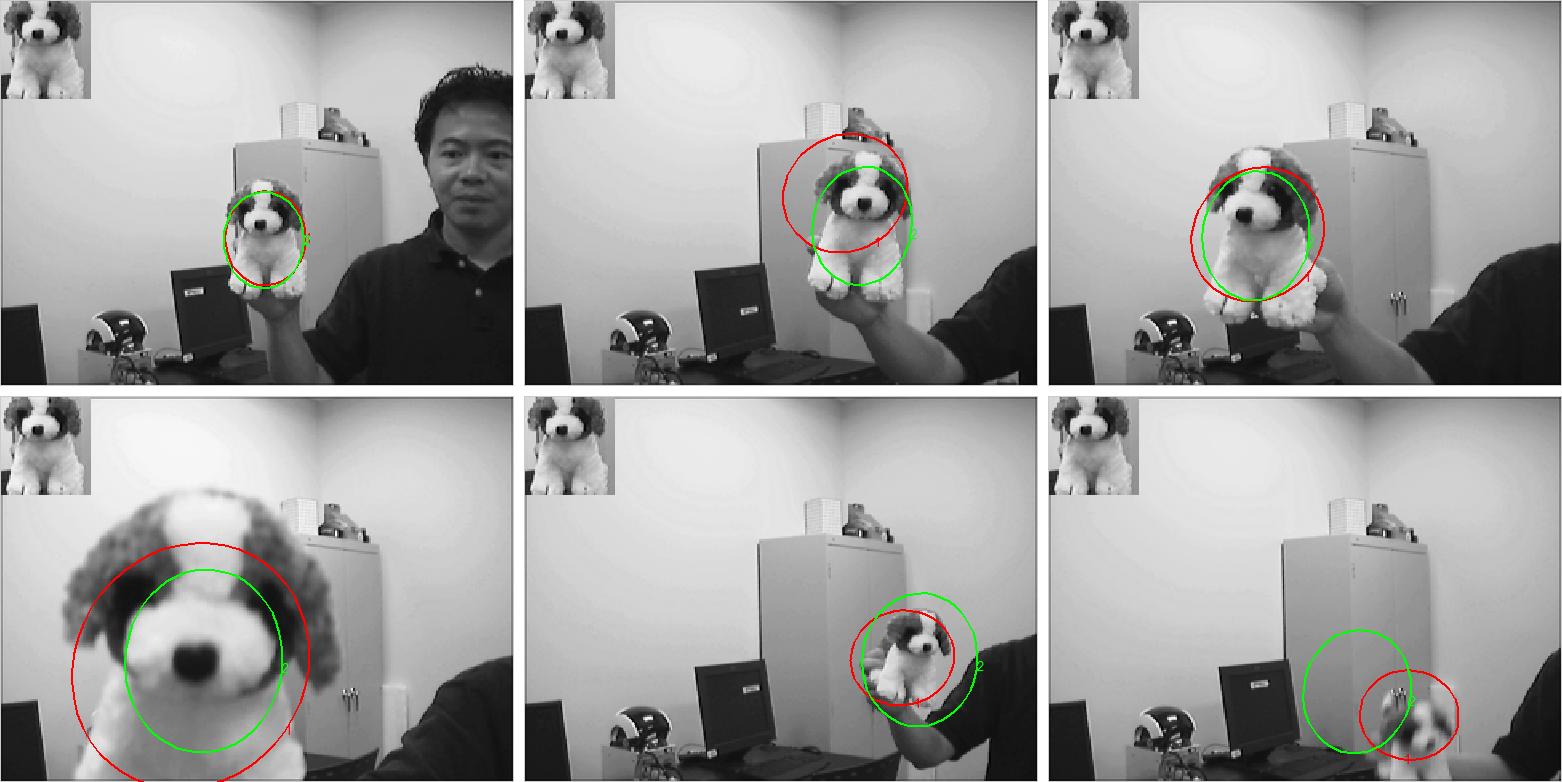}
\caption{Tracking results on toy dog sequences, frame \#1, 450, 715, 1014, 1271, 1351, Green: IPCA, Red: our results. The IPCA was  slightly more localized in stable case, but failed to follow through when the target underwent a fast motion towards frame \#1351.}
\label{fig:res_dog1}\vspace{-10pt}
\end{figure}
\begin{figure}[!t]
\centering
\includegraphics[width=0.48\textwidth]{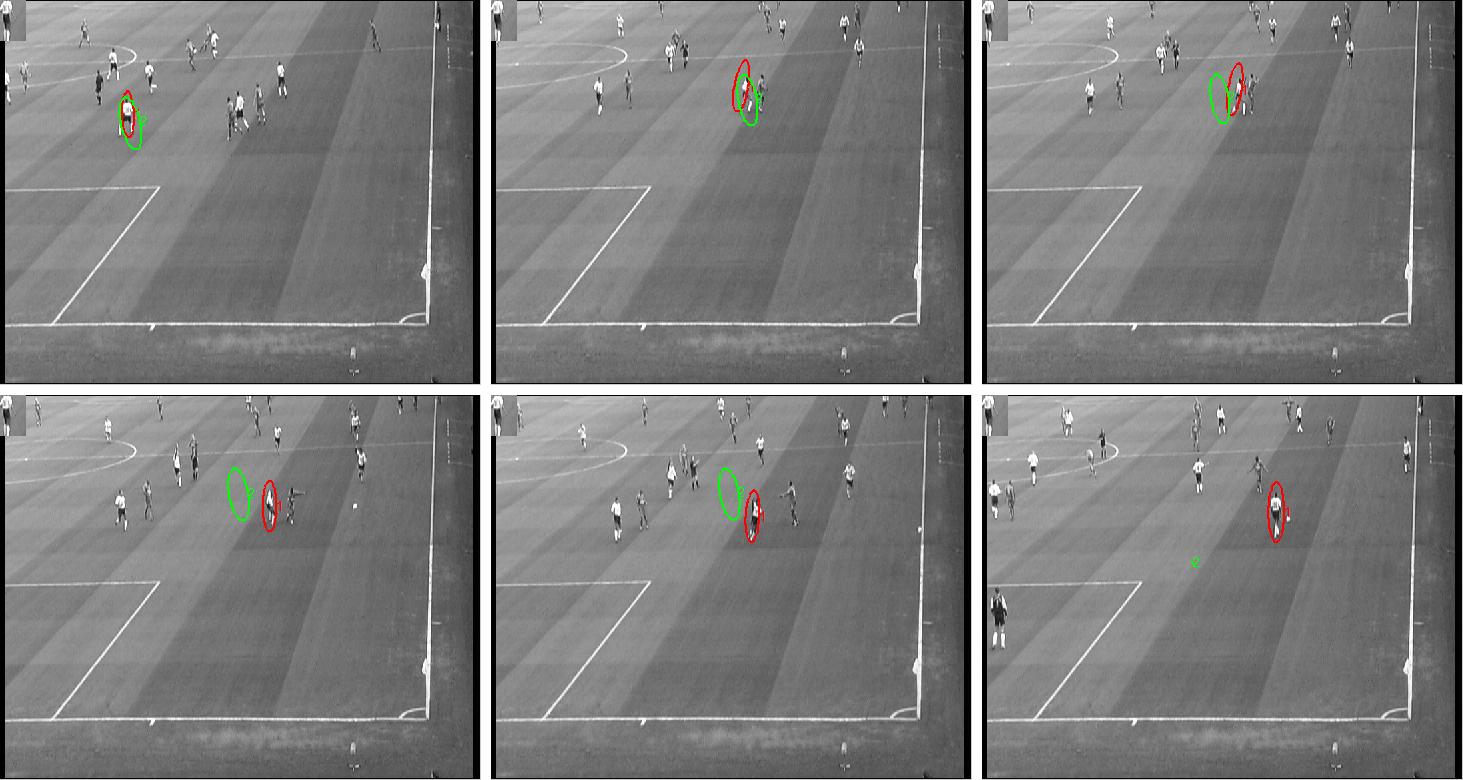}
\caption{Tracking results on soccer sequences, frame \#246, 628, 630, 661, 686, 996, Green: IPCA, Red: our results. The IPCA started to drift off from frame \#628 when the player's legs moved fast, and lost track shortly.}
\label{fig:res_soccer_test7}\vspace{-10pt}
\end{figure}
\begin{figure}[!t]
\centering
\includegraphics[width=0.48\textwidth]{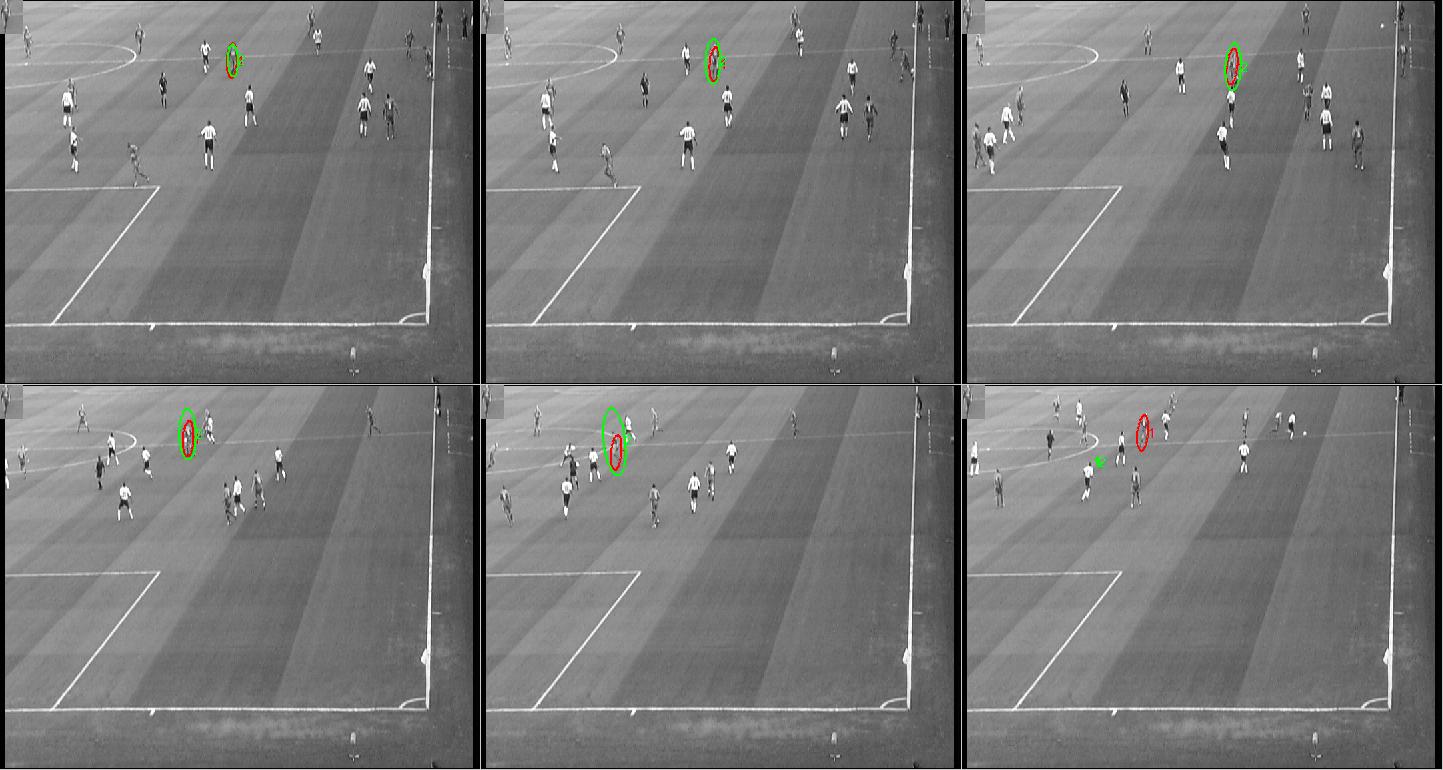}
\caption{Tracking results on soccer sequences, frame \#10, 15, 122, 248, 285, 360, Green: IPCA, Red: our results. The IPCA started to drift off from frame \#15 due to low contrast between the target and the background.}
\label{fig:res_soccer_test10}\vspace{-10pt}
\end{figure}
\begin{figure}[!t]
\centering
\includegraphics[width=0.48\textwidth]{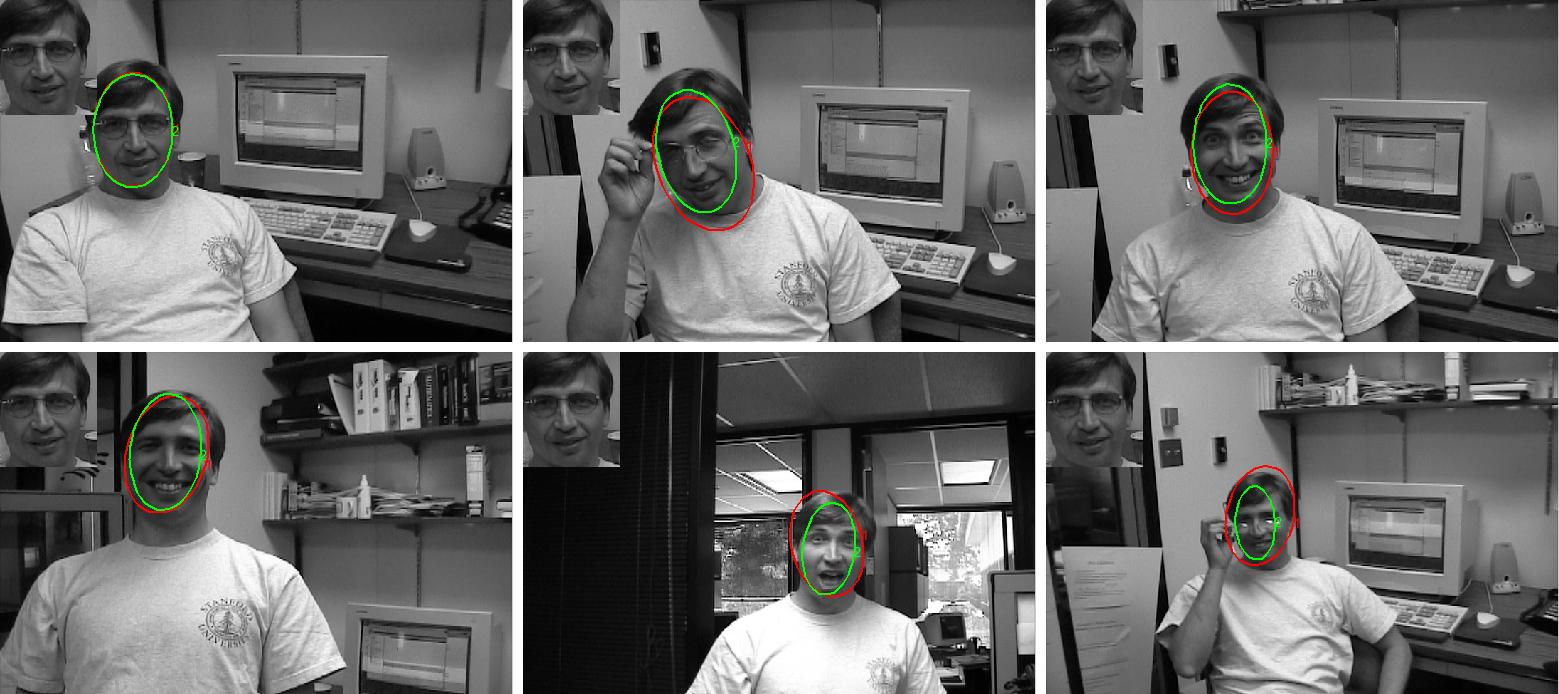}
\caption{Tracking results on DUDEK FACE sequences, frame \#1, 361, 459, 605, 795, 1095, Green: IPCA, Red: our results. Both results were comparable despite of his rich facial expressions, which had more effects on our covariance descriptor.}
\label{fig:res_dudedata}\vspace{-10pt}
\end{figure}
\begin{figure}[!t]
\centering
\includegraphics[width=0.48\textwidth]{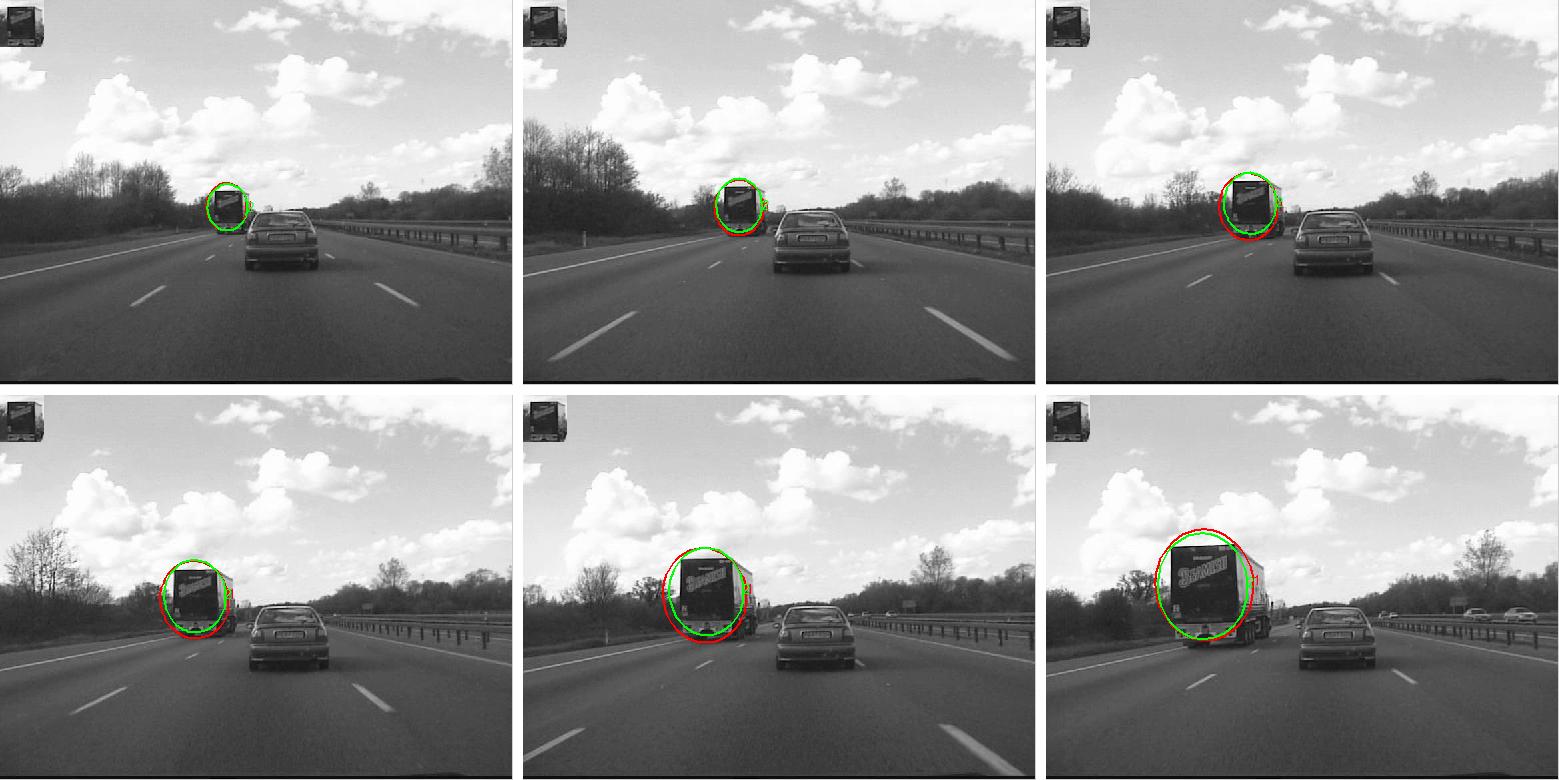}
\caption{Tracking results on PETS2001 vehicle sequences, frame \#1, 25, 50, 75, 100, 125, Green: IPCA, Red: our results.  Both results were comparable.}
\label{fig:res_PETS2001}\vspace{-10pt}
\end{figure}
\begin{figure}[!t]
\centering
\includegraphics[width=0.48\textwidth]{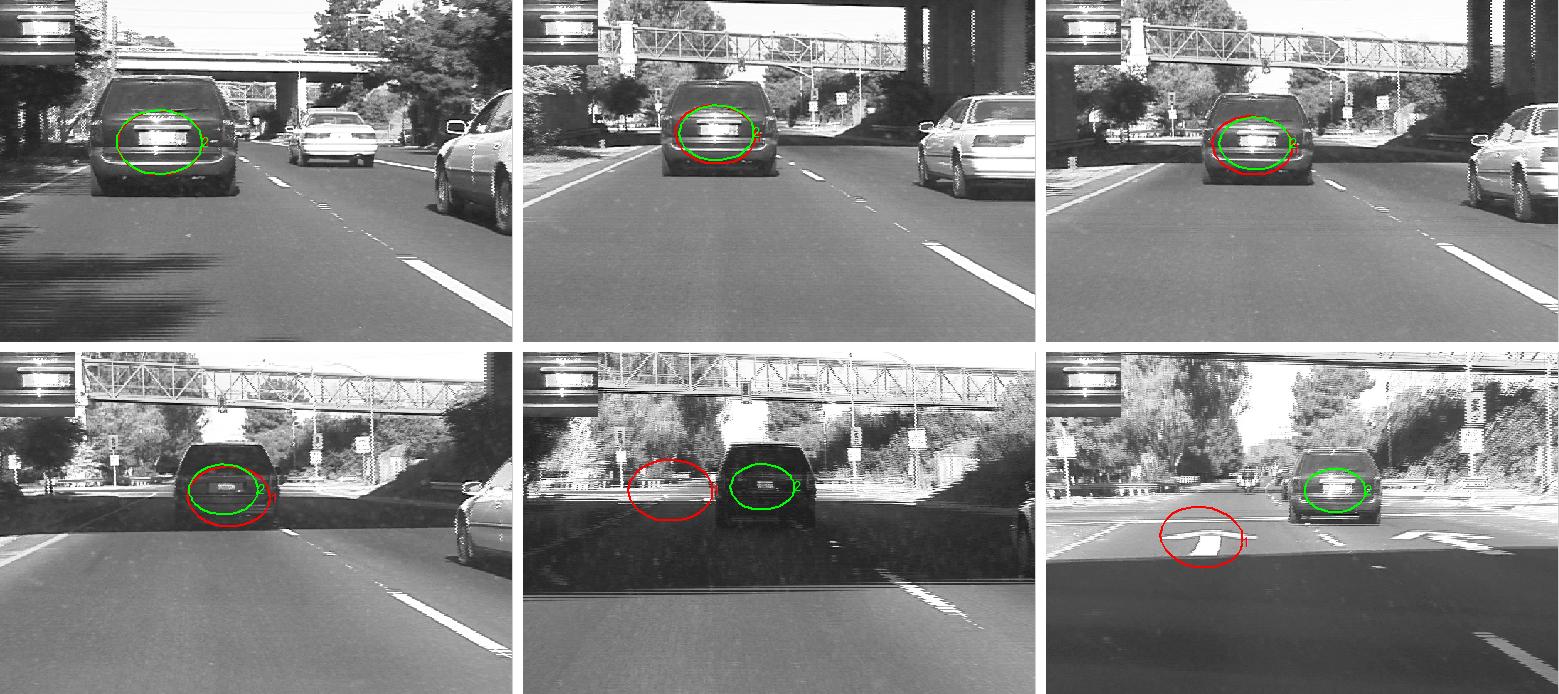}
\caption{Tracking results on car sequences, frame \#1, 132, 150, 168, 184, 227, Green: IPCA, Red: our results. The IPCA performed better, was robust to illumination changes, but our method mainly used template gradients, which changed dramatically due to shadow and lack of reflection of the car plate. At frame \#227, the arrow sign might look too similar to the target in gradients.}\vspace{-10pt}
\label{fig:res_car4}
\end{figure}

%
  We compared our method with the current state-of-the-art algorithm, the incremental PCA (IPCA) method by David et al \cite{Ross08incrementallearning}. Our results are shown in red and the IPCA in green from Figures \ref{fig:res_sylv} to \ref{fig:res_car4}. In PLUSH TOY SYLV sequences shown in Figure \ref{fig:res_sylv},the IPCA failed to recover tracking from frame \#609 when it locked onto the background, which looks more similar to the upright SYLV. Fast poses changes around frame \#609 caused the IPCA eigenbases  non-representative as shown in Figure 4.

  Similarly, in Figure \ref{fig:res_dog1}, the IPCA failed to follow through when target underwent a fast motion towards the frame \#1351. This shortcoming of the IPCA is better reflected in Soccer Sequences of PETS2003. the IPCA started to drift off from frame \#628 shown in Figure \ref{fig:res_soccer_test7} when the player moved his legs fast, and lost track shortly.  In the same sequence in Figure \ref{fig:res_soccer_test10}, the IPCA found it hard to track the opposite team players who wore dark clothes after a short occlusion at frame \#285.

  In Figure \ref{fig:res_dudedata}, Dudek Face sequences, both methods perform well despite of his rich facial expressions, which have more effects on our covariance descriptor.  In the more stable  vehicle sequence from PETS2001 in Figure \ref{fig:res_PETS2001}, again both methods could track well.   Figure  \ref{fig:res_car4}  shows  an example of a car sequence, in which our method did not perform satisfactorily.  Our method  locked onto  the  background  whereas the IPCA showed robustness to the illumination changes. The  possible  explanation  is  that  our template dynamics was unable to account for this dramatic and non-smooth transition of the template when the car went into a shadowed region.  Also, a closer look showed that the IPCA eigenbasis looked similar to the target template in shadows.

The overall tracking performance on the test cases is summarized in Figure \ref{fig:results_stats}. Note that images sequences of Sylv, PETS2001 and soccer player 4 have targets out of the images, this explained the small track duration performance. Nevertheless, our method shown in red generally had longer track length. On the hand, given frames that were on track for both trackers,  IPCA showed better track accuracy shown in Figure \ref{fig:results_stats}b. For the sequences with frequent changes in target appearance such as soccer sequences, the track goodness was comparable. The video sequences may be found on the website, http://www.youtube.com/watch?v=KaSrVbGyvq4.

\noindent\textbf{Discussion}. In stable tracking cases, good pixel-wise alignment enabled the IPCA to track very well. The IPCA was generally very robust to blurring, even illumination changes, as eigenbasis tended to encompass these changes. In other words, some eigenbasis looked similar to blurred or illumination-changed templates.  The distance measure in the IPCA uses a norm of all corresponding pixels difference; as such, it tends to be very stable and well aligned in the stable target cases. On the other hand,  it is likely to favor the relatively stable regions in the target. When such regions are too similar to the background and target poses changes at the same time, then the IPCA may lose track very quickly in the Soccer Sequence in Figure \ref{fig:res_soccer_test10}. On the other hand, our method uses covariance of gradients and intensity; the template feature descriptor is  much smaller in dimension. This may cause our method slightly less precise than the IPCA shown in Figure \ref{fig:res_dog1}, which our method did not match to pixel accuracy. Figure\ref{fig:res_car4}, our method lost track when the vehicle entered the shadowed region, because the both gradients and intensity changed significantly and for an interval.

Although our method was slightly not as precise in the stable cases, it gain much more flexibility in the non-stable tracking scenarios. In the cases of non-rigid or fast motion of targets, mis-alignment in the posterior estimate (the new template sample to add to the eigen space in the IPCA) and eigenbases may accumulate over a short interval and consequently render eigenbases non-representative at all. This inevitably leads to loss in tracking.  Our method could deal with these scenarios a lot better for two reasons. Firstly, the template descriptor did not require pixel-wise alignment and is robust to mis-alignment. Secondly, the generative process could accommodate multiple hypothesis of the template on the covariance Riemannian manifold, and it automatically selects the better hypothesis as the target template evolves as shown in Figure \ref{fig:random walk on the manifold}.

However, there are some limitations in our algorithm. One of them is to the need to careful choose a suitable region for tracking. Since we used the published features such as intensity and gradients, and second order gradients for covariance, these features are sensitive to specular effects, dark shadows as shown in Figure \ref{fig:res_car4}. It is also important to choose a target region with fairly good gradients variations, otherwise the covariance descriptor may be ill-conditioned consequently affecting both eigenvalues estimation and distance measurements in Equation \eqref{eq:distCov}.
\section{Conclusion}
\label{section:conclusion}
In this paper, we have proposed a new method to update
target model in tandem with the target kinematics. More
precisely, we have developed a generative template model in a
principled way within a Bayesian framework. A novel template propagation mechanism in the log-transformed space of the covariance manifold to free the constraints inherently imposed by positive definite matrices.  We have shown that the simple generative process can allow template to evolve naturally with target appearance variation. It is hoped that
by jointly quantifying the uncertainties of the target kinematics
and template, we are able to achieve more robust visual tracking. We have chosen the covariance descriptor as the target representation. We have modeled the target template model
dynamic using a random walk on the covariance Riemannian manifold. Our template dynamic model is an example of a diffusion process on the covariance Riemannian manifold.
In the experiments, our algorithm outperformed with the current state-of-the-art algorithm IPCA particularly when the target underwent a fast and non-rigid poses changes, and also maintained a comparable performance when the target was more stable. Some future work includes automatic selection of covariance features that are  more robust to a sudden dramatic change in illuminations.

 Future work includes addressing a number of questions such as how should the diffusion speed be adjusted and can the diffusion process be better constrained. Another area of work is to deal with illumination changes in the manifold generative process. In order to improve the goodness of track, a  more discriminative target descriptor is to be explored.

%
%
%
%
%
%
%


%

%

\ifCLASSOPTIONcompsoc
  \section*{Acknowledgments}
\else
  \section*{Acknowledgment}
\fi
We would like to thank DSO National Laboratories, Singapore for partially sponsoring this work, and David Ross for sharing the test sequences.

\ifCLASSOPTIONcaptionsoff
  \newpage
\fi

\nocite{*}



%

{
\bibliographystyle{ieee}
\bibliography{egbib}
}
\begin{IEEEbiographynophoto}{Marcus Chen} is a PhD student in the School of
Computer Engineering, Nanyang Technological University, and member of technical
staff in DSO National Laboratories. He received his B.S. in Electrical and Computer Engineering with  university honours in 2007 from Carnegie Mellon University, and M.S. in Eletrical Engineering from Stanford University in 2008.
\end{IEEEbiographynophoto}
\begin{IEEEbiographynophoto}{Dr. Cham Tat Jen} is an Associate Professor in the School of Computer Engineering, Nanyang Technological University, and Director of the Centre for Multimedia \& Network Technology (CeMNet). He received his BA in Engineering with triple first class honours in 1993 and his PhD in 1996, both from the University of Cambridge, during which he was awarded the Loke Cheng-Kim Foundation Scholarship, the Alexandria Prize, the Engineering Members' Prize as well as the St Catharine's College Senior and Research Scholarships. Tat-Jen was subsequently conferred a Jesus College Research Fellowship in Science in 1996-97. From 1998 to 2001, he was a research scientist at DEC/Compaq CRL in Cambridge, MA, USA, where his experience included technology transfer to product groups and showcasing research work to Hollywood studios. After joining NTU in 2002, he was concurrently a Faculty Fellow in the Singapore-MIT Alliance Computer Science Program in 2003-2006.
\end{IEEEbiographynophoto}
\begin{IEEEbiographynophoto}{Dr. Pang Sze Kim} is a principal member of technical staff and currently is the head of Signal Processing Laboratorie at DSO National Laboratories. Sze Kim obtained his Ph.D. from University of Cambridge.
\end{IEEEbiographynophoto}
\begin{IEEEbiographynophoto}{DR. Alvina Goh}obtained her Ph.D. from the Department of Biomedical Engineering at the Johns Hopkins University in May 2010. Currently, she is a researcher at the DSO National Labs and an adjunct assistant professor in the Department of Mathematics at the National University of Singapore. Her research interests include machine learning, computer vision, and medical imaging.
\end{IEEEbiographynophoto}



\end{document}